\def\year{2018}\relax
\documentclass[letterpaper]{article}
\usepackage{aaai18}
\usepackage{times}
\usepackage{helvet}
\usepackage{courier}

\usepackage{amsmath}
\usepackage{amssymb}
\usepackage{algorithm}
\usepackage{algpseudocode}
\makeatletter
\def\BState{\State\hskip-\ALG@thistlm}
\makeatother
\usepackage[caption=false]{subfig}

\usepackage{mathtools}

\usepackage{url}  
\usepackage{graphicx}  
\usepackage{bm}
\usepackage{multirow}
\algrenewcommand\algorithmicrequire{\textbf{Input:}}
\algrenewcommand\algorithmicensure{\textbf{Output:}}

\algnewcommand\algorithmicforeach{\textbf{for each}}
\algdef{S}[FOR]{ForEach}[1]{\algorithmicforeach\ #1\ \algorithmicdo}

\usepackage{booktabs}

\usepackage{color}

\begin{document}
%
\title{Topical Phrase Extraction from Clinical Reports by Incorporating both Local and Global Context}
\author{Gabriele Pergola, Yulan He \and David Lowe\\
 School of Engineering and Applied Science, Aston University, UK\\
  {\tt pergolag@aston.ac.uk, y.he@cantab.net, d.lowe@aston.ac.uk}
}

\maketitle
\begin{abstract}
\begin{quote}

Making sense of words often requires to simultaneously examine the surrounding context of a term as well as the global themes characterizing the overall corpus. Several topic models have already exploited word embeddings to recognize local context, however, it has been weakly combined with the global context during the topic inference. 
This paper proposes to extract topical phrases corroborating the word embedding information with the global context detected by Latent Semantic Analysis, and then combine them by means of the P\'{o}lya urn model. To highlight the effectiveness of this combined approach the model was assessed analyzing clinical reports, a challenging scenario characterized by technical jargon and a limited word statistics available. Results show it outperforms the state-of-the-art approaches in terms of both topic coherence and computational cost.

\end{quote}
\end{abstract}

\section{Introduction}
Topic models have been extensively used to generate synthetic representations of the main themes characterizing a large document collection. Documents are traditionally represented under the bag-of-words assumption, a simple but effective representation that ignores the word orders, but in spite of this has shown worth noting results \cite{Blei03}.
However, this assumption has commonly led to the extraction of unigram topics relying on the word co-occurrence patterns across documents. This has notably narrowed the topic expressiveness as the shared semantic of words is solely based on the global context; moreover, many domain-specific documents might include concepts that are unfolded in multiple words rather than in a single term.
Clinical reports are a prominent example of this family as medical concepts are often expressed in terms of multi-word phrases. For example, the phrases ``\textit{white blood cell}" or ``\textit{blood sugar}" would lose their meaning if decomposed as unigrams; in addition, the word \textit{cell} and \textit{sugar} might be wrongly put under the same topic because of the shared \textit{blood} term.

Recently, word embeddings have gained an increasing interest thanks to their improved efficiency in representing words as continuous vectors of a low-dimensional space \cite{Mikolov13a,Joulin16}. The resulting embeddings have been proved to encode numerous semantic relations (e.g. similarity or analogies) based on the local context of words \cite{Levy15}. Several works have combined topic models with word embedding \cite{nguyen2015improving,Li16}. This has enhanced the semantic coherence of the topics discovered since words with similar semantic and syntactic properties are close to each other in the embedding space which overcomes the curse of dimensionality when representing words as atomic units. However, these models commonly entail a significant computational cost and are exposed to the \textit{topic shifting} problem \cite{Rekabsaz17}. 
Indeed, words that share similar context windows might potentially be treated as directly co-related into the embedding space, misleading word similarity in case of antonyms (e.g. \textit{tall} and \textit{short}) or co-hyponyms (e.g. \textit{schizophrenia} and \textit{alzheimer}). 

The computational cost required to combine word embeddings and topic models can be reduced by adopting the \textit{Generalized P\'{o}lya urn model} \cite{Mahmoud08}. Although the Latent Dirichlet Allocation (LDA) \cite{Blei03} already used the \textit{Simple P\'{o}lya urn model}, its generalized version proposed in Mimno et al. (\citeyear{Mimno11}) allows to incorporate word relatedness directly into the inference process. Hence, a simple extension would be evaluating word relatedness based on word embeddings for the \textit{Generalized P\'{o}lya urn model}. 
The topic shifting problem can be mitigated by jointly considering the global and local context of a word: if two terms appear in similar context-windows but do not share similar global contexts (i.e. corpus themes), they probably convey different topics. 

In this paper, we propose a Context-aware P\'{o}lya urn model (Context-GPU) to generate topics by extracting topical phrases combining the local and global context of words/phrases\footnote{https://github.com/gabrer/context\_gpu/}. We first detect the medical phrases in clinical reports by means of an off-the-shelf medical concept extraction tool; hence, the phrases extracted are thus reliable and clinically relevant. Then we use a modified Generalized P\'{o}lya urn model, which promotes words/phrases under the same topic if they are close neighbors in the window-based embedding (local context) as well as in the corpus-based embedding (global context) space. The window-based embedding improves the capability to detect semantic relatedness at the phrase level; also, it encodes word co-occurrences from an external source of knowledge (e.g. Wikipedia) alleviating the lack of statistics for technical terms. Simultaneously, the corpus-based embedding limits the topic shifting during topic inference. To the best of our knowledge, this is the first time local and global context are combined for topical phrases extraction. Our experimental results have shown the effectiveness of this approach outperforming the previous methods in terms of quality of topics, topic coherence and efficiency.

We proceed to describe the related work. We then give a background of the P\'{o}lya urn model before presenting the proposed approach. Finally, we discuss our experimental results in comparison with the state-of-the-art approaches to topical phrase extraction.

\section{Related work}

Our work is related to three lines of research, phrase embedding learning, topic modeling incorporating word embeddings and using latent topics for language model learning.

\subsection{Phrase Embedding Learning} 

Distributional semantic models (i.e. word embeddings) have recently been applied successfully in many NLP tasks \cite{Levy15}. Neural network based approaches have become more efficient, allowing their use in multiple scenarios, thanks to the \textit{skip-gram with negative-sampling training method} (SGNS), \cite{Mikolov13a,Mikolov13b}. It was widely popularized via \textit{word2vec}, a software to create word embeddings. Recently, a new word embedding method has been proposed, called \textit{FastText} \cite{Joulin16}, which treats each word as made of character n-grams. Vector representations are then computed from the sum of their n-gram representations.
More traditional vector representations are based on a dimensionality reduction obtained by applying the Singular Value Decomposition (SVD) to the weighted document-term matrix of the corpus; Latent Semantic Analysis (LSA) \cite{Deerwester90} is a prominent method following this approach.

Phase embeddings can be simply taken as the average of their constituent word embeddings. If treating each phrase as a single term, its representation can also be learned from data directly using word representation learning methods such as LSA, SGNS or FastText. There have also been compositional semantic models which aim to build distributional representations of a phrase from its constituent 
word representations using Convolutional Neural Networks (CNNs) \cite{le2014distributed}, based on features that capture phrase structure and context \cite{yu2015learning} or using convolutional tensor decomposition \cite{huang2016unsupervised}.

\subsection{Topic Modeling Incorporating Word Embeddings}

To exploit the information encoded into word embeddings, several models have been proposed combining topic models and word embedding representations. Gaussian LDA \cite{Das15}, for instance, use pre-trained word embeddings learned from large corpora (e.g., Wikipedia) to model topics as Gaussian distributions over the vector representations, defining topics as random samples from a multivariate Gaussian distribution whose mean is the topic embedding. 

Nguyen et al. (\citeyear{nguyen2015improving}) proposed to use the word embeddings pre-trained from an external large corpus as latent word features to define categorical distributions over words, which is called a latent feature component. The original topic-to-word Dirichlet multinomial component in LDA which generates the words from topics is then replaced by a two-component mixture of the original Dirichlet multinomial component and a latent feature component. But model learning is difficult because of the coupling between the two components.

An alternative approach is TopicVec \cite{Li16} which replaces the multinomial topic-word distribution with a probability function, it computes a focus word from a topic and word neighbors within the embedding; in TopicVec this link function is in addition combined with a context word embeddings along with the topic embedding and the focus word embedding.

Li et al. (\citeyear{Li16SIGIR}) measured the word relatedness based on pre-trained word embeddings and used it to modify the Gibbs sampling inference in a generalized P\'{o}lya urn model; overall, this strategy significantly reduces the computational cost compared to the aforementioned approaches. However, not only it is fully focused on the short-text analysis (i.e. one document one topic), but it doesn't exploit any global context to mitigate the topic shifting issues induced by word embeddings. Also, it didn't explore the benefit of using a $n$-gram word embedding such as FastText against the word-oriented embeddings.

\subsection{Using Latent Topics for Language Model Learning}

While the aforementioned approaches incorporate the word embeddings into topic model learning, there have also been attempts making use of latent topics to improve language models. Dieng et al. (\citeyear{dieng2017topicrnn}) proposed TopicRNN in which the global semantics come from latent topics as in typical topic modeling, but local semantics is defined by the language model constructed using Recurrent Neural Networks (RNNs). The separation of global vs local semantics is achieved using a binary decision model for stop words. Topic vectors here are also sampled from a Gaussian distribution with zero mean and unit variance and are refined during language model learning. In a similar vein, Lau et al. (\citeyear{lau2017topically}) proposed a topic-driven neural language model that also incorporates document context in the form of latent topics into a language model implemented using Long Short-Term Memory (LSTM) networks. They essentially treated the language and topic models as subtasks in a multi-task learning setting, and trained them jointly using categorical cross-entropy loss.

\section{P\'{o}lya urn Models}
In this section, we give a background of both simple and generalized P\'{o}lya urn Models. We describe how they can be used for topic extraction, before presenting in the next section, our proposed approach that extends them to exploit word contexts. 

As shown in Mimno et al. (\citeyear{Mimno11}), simple LDA model might not be able to fully capture the already available statistics of word co-occurrences in a corpus. Detecting semantic similarity between words is challenging due to the power-law characterization of natural language, i.e., words sharing a common semantic might rarely co-occur together and hence being overlooked. 
A more effective model called Generalized P\'{o}lya urn model was proposed in Mimno et al. (\citeyear{Mimno11}), by extending the Simple P\'{o}lya urn model used in LDA where the topic-word component is updated in order to strengthen the associations between related words under the same topic. 


\subsection{Simple P\'{o}lya Urn Model}
The generative process of LDA can be interpreted by means of P\'{o}lya urn model \cite{Mahmoud08}, a statistical model describing objects of interest (e.g. words or topics) in terms of colored balls and urns. 

In the context of topic models, balls can be considered as words and urns as topics; in particular, LDA follows the so-called \textit{Simple P\'{o}lya urn} (SPU) model. In the main step of this process, a colored ball is randomly drawn from an urn and is put back along with an additional new ball of the same color; this induces a self-reinforcement process known as "rich get richer", since the probability of seeing a specific colored ball from an urn increases every time this ball has been drawn.

Likewise, LDA follows the SPU model by employing two kinds of urns: topic-document and word-topic urns. The topic-document urns hold balls whose color corresponds to different topics in a document, while the balls in the word-topic urns represent different words in a topic. 
The generative process proceeds as follows: a ball is extracted from the topic-document urn $d_m$, and its color determines the new topic assignment $\hat{z}$, then the ball is put back along with another ball of the same color. Next, a ball is extracted from the word-topic urn $\hat{z}$ determining a new word $\hat{w}$ and, as before, the ball with an additional one of the same color is put back into the urn.
As a result, both the topic $\hat{z}$ and the word $\hat{w}$ increase their proportion in the topic-document and word-topic distribution, respectively.

\subsection{Generalized P\'{o}lya Urn Model}
The described process is intrinsically biased to promote together words that frequently occur in a corpus, overlooking less prominent but correlated words. To alleviate this shortcoming and increase the association strength between rare but still related words a \textit{Generalized P\'{o}lya Urn} (GPU) model was proposed by  Mimno et al. (\citeyear{Mimno11}). It incorporates a corpus-specific word co-occurrence metric into the generative process affecting the probabilities of related words under the same topic.

Unlike the aforementioned simple version, in a generalized P\'{o}lya urn model when a ball of color $\hat{w}$ has been drawn, $A_{vw}$ additional balls of several colors $v=\{1,...,W\}$ are placed into the urn. This process increases, not only the probability of the observed word $\hat{w}$, but also the probability of its related words, and is commonly referred as \textit{promotion} of the colored balls \cite{Fei14}.
Specifically, the LDA inference process now relies on a modified Gibbs sampling algorithm which simultaneously increases the probability of a word and their correlated terms at each iteration. Word relatedness is computed by weighting word co-occurrences using the standard Inverse-Document Frequency (IDF) weighting strategy $\lambda_v = log(D/D(v))$, where \textit{D} is the number of documents and \textit{D(v)} is the number of document where the word \textit{v} occurs at least once; this weight has the beneficial property of being higher for rare words increasing their prominence. 

However, the effectiveness of this approach strongly depends on how accurately word correlations are identified.
Although the GPU framework proposed by  Mimno et al. (\citeyear{Mimno11}) has improved the average quality of mined topics, it still relies exclusively on the global context of words (i.e. word co-occurrences in the corpus) and might completely overlook sentence-specific meaning of a word conveyed by the word's local context.

This drastically narrows the model capability to deal with multiple sense of words. For example, looking at the sentences ``\emph{White blood cell count is low.}" and ``\emph{This raises the blood sugar back to its normal level.}", current model might put under the same topic words like ``\emph{cell}" and ``\emph{sugar}", which are rather unlikely to appear coupled in a sentence. Moreover, similar issues can be experienced analyzing documents characterized by technical jargons which occur few times in corpus (i.e. poor statistics) and might exhibit a peculiar meaning for every phrase (i.e. multiple meanings).

\section{Context-Aware P\'{o}lya Urn  (Context-GPU) Model}

In this section, we propose a modified Gibbs sampling algorithm to conduct a context-driven inference to cope with the described limitations. It exploits a word representation based on general knowledge source which provides a rich word statistics, and takes into account simultaneously the local and global context of words to disambiguate topically-irrelevant terms. 

Our hypothesis is that the Generalized P\'{o}lya Urn model can be modified and enhanced to provide a framework combing local and global context of words. Local context is determined by a word embedding based on context window and trained on a large source of general knowledge (e.g. Wikipedia). 
Rather, the global context lies on the word representations obtained considering the term co-occurrences within a corpus. As a result, both local and global context can be incorporated into a context-aware P\'{o}lya urn model called \textit{Context-GPU}, a generative model which is able to capture the semantics of a word with regard to both sentence and document context, mitigating the topic shifting issue induced by word embeddings. 

Before presenting our proposed Context-GPU, we first describe how we extract medical phrases from clinical documents.


\subsection{Medical Phrase Extraction}

Medical terms in clinical documents are often expressed in multi-word phrases, for example, ``\textit{arterial blood gas}" and ``\textit{heart transplant}". These phrases are not semantically decomposable, as once split into unigrams, they would lose their original semantic meanings.

We use an open source clinical annotation tool \textit{MedTagger}\footnote{http://ohnlp.org/index.php/MedTagger} which extracts and annotates concepts from clinical reports by leveraging knowledge bases, machine learning and syntactic parsing. The output of MedTagger provides detailed information about the medical concept detected such as attributes, uncertainty, semantic group (i.e. Diagnosis, Test and Treatment) and so on. Also, it has achieved the-state-of-the-art performance in terms of F-Measure (0.84) at the \textit{i2b2 NLP challenge} on the concept mention task \cite{liu12}. 

Clinical reports are also characterized by many occurrences of medical abbreviations to favor brevity due to a large amount of information that needs to be synthesized in a short time and limited space. Detection of medical concepts through MedTagger is not only much more reliable than other general purpose techniques for phrase extraction, but allows also to effectively detect and preserve the medical abbreviations. 
Once medical phrases are detected, they are represented by compound words where constituent words are joined together by an underscore. E.g. words that compose the phrase ``\textit{short of breath}" are substituted by the compound word ``\textit{short\_of\_breath}", ``\textit{saphenous vein graft}" by ``\textit{saphenous\_vein\_graft}", and so on.  

However, treating phrases as compound words leads to more severe data sparsity since phrases sharing common lower-order $n$-grams, such as ``\emph{right coronary artery}" and ``\emph{left coronary artery}" would be considered as two totally different terms. 
Also, preserving both the multiple words and the compound phrase is not a solution, as the phrases are naturally less frequent than individual words and would be ranked with lower probabilities.

To this end, once multiple words are substituted by a compound word, in the Context-GPU we adopt the FastText embedding \cite{Joulin16}, a word embedding oriented by design to deal with sub-grams composing words. Thus, it naturally fits the need to detect the similarity between a phrase and its constituent words. For example, in our trained FastText embeddings the word \textit{saphenous\_vein\_graft} has neighbors such as \textit{saph}, \textit{aphe}, but also \textit{vein} and \textit{graft}. 
Therefore, we combine FastText with the P\'{o}lya urn model to increase the probability to see under the same topic the words \textit{vein} or \textit{graft} once we come across the phrase \textit{saphenous vein graft}, and vice versa.

\subsection{Local Context}
Some commonly used embedding have been the SVD \cite{Levy15} and SGNS (i.e. word2vec) \cite{Mikolov13a}, and only recently FastText \cite{Joulin16}, and they all provide vectors encoding both syntactic and semantic information about a word at its local context window in a large corpus.

Two characteristic features of these embeddings are here exploited. The first is that words are represented by a vector trained with regard to the local contexts where the words are likely to appear. Therefore, it can be used to reinforce ties among words sharing common uses in phrases (e.g. \textit{alzheimer} and \textit{schizophrenia}). The second feature is that word embeddings are commonly trained on a large external source of data (e.g. Wikipedia), hence they can mitigate the low statistics of infrequent technical jargons or rare words in a corpus. 

Words are considered related based on the geometric proximity of their vector representations. We propose two strategies to extract related words: a threshold and a Top-$N$ approach. For the threshold approach, the related words of a given word are those whose cosine distances with the target word are less than a pre-defined threshold. 
Alternatively, the Top-N approach extracts a fixed number $N$ of the closest words regardless of their actual distances. 
In the former approach, the number of neighbors is not fixed for different words, while in the latter the number is fixed but also unrelated words could be added to the neighbor set. 



\subsection{Global Context}

Although topics extracted by combining word embedding and P\'{o}lya urn model are more consistent with the occurrence pattern of words in sentences, word embeddings have some well-known shortcomings related to antonyms (e.g. \textit{tall} and \textit{short}) or co-hyponyms (e.g. \textit{schizophrenia} and \textit{alzheimer}). Indeed, these are words that might share a similar context windows and then be potentially treated as directly correlated into the embedding space. To avoid any topic shifting resulting from word ambiguities, 
we balance the local context information with the corpus-specific context computed by applying the Latent Semantic Analysis (LSA) \cite{Deerwester90}.

In particular, we use LSA to learn latent topics from data by performing Singular Value Decomposition (SVD) on the $V\times D$ term-document count matrix where $V$ is the vocabulary size and $D$ is the number of documents. SVD factorizes such a matrix into the product of three matrices, $W, \Sigma$, and $C^\intercal$ . In $W\in\mathbb{R}^{V\times m}$, each row represents a word and each column represents a dimension in a latent space which is orthogonal to each other. $\Sigma$ is a diagonal $m\times m$ matrix which contains singular values along the diagonal indicating how important each latent dimension is. In $C^\intercal\in\mathbb{R}^{m\times D}$, each row represents one of the latent dimensions and each column represents a document. If taking the top $k$ latent dimensions in $W$, we will have a reduced matrix $W_k\in\mathbb{R}^{V\times k}$ where each word is essentially represented by a dense $k$-dimensional vector. Hence, using LSA, we will be able to generate another set of word embeddings based on global context. For each word, we can then retrieve its related words using the thresholding or Top-$N$ approaches mentioned above.

One may argue that topic models such as LDA already captures the global context information by compressing the original document into a lower-dimensional bag-of-topics representation. It is worth noting that LSA learns latent topics by performing SVD on term-document count matrix, and as a result, the topics are assumed to be orthogonal. LDA uses generative probabilistic models to generate latent topics which are represented as word distributions, and it uses Dirichlet priors for both the document-topic and topic-word distributions. In LDA, topics are allowed to be non-orthogonal. So although both LSA and LDA try to capture the global context, the topic results would be somewhat different. It has been pointed out previously that in some scenarios LSA outperforms LDA providing better quality topics \cite{Bergamaschi15}. As will be shown in our experiments, additionally incorporating the global context derived by LSA into the context-aware Polya Urn model gives better performance.

Words are likely to express a common topic, not only when sharing a common local context window (e.g. FastText similarity), but also a global context (i.e. LSA similarity) depending on the analyzed documents.
Therefore, we first extract the word neighbors both from local context and global context based embeddings, and we then preserve only the terms in the intersection of those sets, hence improving the probability that a word in a topic sharing both local and global context.

\subsection{Topic Inference}

Given the set of documents $\mathcal{D}$ and the topic assignments $\mathcal{Z}$, the conditional posterior probability of a word $w$ in a topic $z$ follows the standard generalized P\'{o}lya urn model \cite{Mimno11}:

\begin{equation}
  P(w |z,\mathcal{W},\mathcal{Z},\beta,\mathbf{A}) = \frac{\sum_{v} N_{v|z} A_{vw} + \beta}{N_z + |\mathcal{V}|\beta}
\label{eq:gpumodel}
\end{equation}

where $\mathbf{A}$ is a \textit{promotion matrix} that expresses whether two words are related to each other, i.e., if one should influence the expectation to draw the other one.

The promotion matrix is critical for the overall algorithm performance, as it concisely expresses the available information about word relatedness. We propose to set the values of $\mathbf{A}$ by computing the word relatedness as a result of the $\mathcal{P}$ neighbors provided by the local context embeddings and the $\mathcal{Q}$ neighbors from the global context embeddings. 
For a word \textit{v}, another word \textit{w} is promoted if it is \textit{v}'s neighbor both at the local level (i.e., based on its local context embedding) and the global level (i.e., based on its global context embedding), as expressed in Eq. \ref{eq:matrixA}. Thus, only if both words are correlated in both the local and global context embedding space, their corresponding cell value in $\mathbf{A}$ is updated to increase their probabilities to be drawn under the same topic. 
In the particular case of $\mathbf{A}$ being the identity matrix, the model collapse into the the Simple P\'{o}lya urn model, providing the posterior probability of a word \textit{w} under a topic \textit{z} in standard LDA.

\begin{algorithm}[htb]
\begin{algorithmic}[1]
\caption{Training procedure of the context-aware P\'{o}lya urn model.}
\label{alg:polyaurn_model}
\Require Corpus C, K topics, $\alpha$, $\beta$, thresholds $\tau$ and $\sigma$
\Ensure Posterior topic-word distribution

\State \texttt{/* Medical phrase extraction */}
\State $C_p \leftarrow MedTagger.PhraseDetection(C);$ \\

\State  \texttt{/* Local and global neighbors */} 
\For{$v \in \mathcal{W}$}
	\State $\mathcal{P}_{v}\, \leftarrow  WindowEmbedding.Neighbors(v)$;
    \State $\mathcal{Q}_{v} \leftarrow  CorpusEmbedding.Neighbors(v)$;
\EndFor \\

\State \texttt{/* Promotion matrix */} 
\State $\mathit{A_{v,w}} \leftarrow ComputePromotionMatrix(\mathcal{P}_{v}, \mathcal{Q}_{v})$ \\

\State \texttt{/* Generalized P\'{o}lya Urn sampling */} 
\For{$d \in \mathcal{D}$}

    \For{${w_n} \in \textbf{\textit{w}}^{d}$}
        \State $N_{z_i|d_i} \leftarrow N_{z_i|d_i} - 1 $
        \For{\textbf{all} $v$}			
        	\State $N_{v|z_i} \leftarrow N_{v|z_i} -$ \textbf{\textit{$A_{vw_{i}}$}}
    	\EndFor
    \EndFor
    
	\State \texttt{sample} $z_i \propto (N_{z|d_i} + \alpha_z) \frac{N_{w_i|z} + \beta}{\sum_{z'}^{} N_{w_i|z'} + \beta}$
    
    \For{${w_n} \in \textbf{\textit{w}}^{d}$}
        \State $N_{z_i|d_i} \leftarrow N_{z_i|d_i} + 1 $
        \For{\textbf{all} $v$} 			
        	\State $N_{v|z_i} \leftarrow N_{v|z_i} +$ \textbf{\textit{$A_{vw_{i}}$}}  
    	\EndFor
    \EndFor
\EndFor

\end{algorithmic}
\end{algorithm}

The training procedure of our proposed context-aware P\'{o}lya urn model is shown in Algorithm \ref{alg:polyaurn_model}. 
The Gibbs sampling inference can be more complex and expensive due to the non-exchangeability property of words in the generalized P\'{o}lya urn model (i.e. under the same topic, joint probability of words is not invariant to permutation). Therefore, we follow the same approach adopted in Mimno et al. (\citeyear{Mimno11}), considering each word as it was the last one during the inference process, ignoring the effect for subsequent words and their topic assignments. 

\begin{equation}
  \mathit{A_{v,w}}= 
  \begin{dcases}
      1 & \text{if  }  w \in (\mathcal{P}_{v} \cap \mathcal{Q}_{v})  \\
      0               & \text{otherwise}
  \end{dcases}
\label{eq:matrixA}
\end{equation}




\section{Experiments}

We assess the effectiveness of our proposed Context-GPU using the data released as part of the i2b2 Natural Language Processing Challenges for Clinical Records \cite{Uzuner10}. The corpus consists of 1,243 de-identified discharge summaries, characterized by medical jargon, which describes medications, dosages, modes (e.g. oral, intravenous, etc.), frequencies, reasons for the treatment, and so on. Hence, we adopted this dataset to assess the model efficacy to deal with multi-phrase concepts and domain-specific jargon. 

Clinical reports are preprocessed by removing the English common stop words as well as the clinical stop words (e.g. "Dr.", "medical problem", "discharge", etc.). We filter out the most frequent 10 words and the words occurring less than 5 times. We use the MedTagger software to detect the medical phrases and substitute them within the documents. We do not perform stemming. As a result, in the ``bag-of-words" representation the vocabulary size is 7,883, while in the ``bag-of-phrases" increases to 9,932.

Word embeddings are trained on a Wikipedia 2015 snapshot combined with the i2b2 dataset. We use the \textit{hyperwords} library\footnote{https://bitbucket.org/omerlevy/hyperwords} \cite{Levy15} to train the 300-dimensional SVD and SGNS embeddings, configured with the default parameters. Likewise, we train the 300-dimensional FastText embedding using the library provided by Facebook Research\footnote{https://github.com/facebookresearch/fastText} and setting the n-gram sizes between 2 and 6.
The LSA representation adopted as local context is computed on the i2b2 dataset; we use the S-space library\footnote{https://github.com/fozziethebeat/S-Space} to obtain the final 300-dimensional representation of words and documents.

We train Context-GPU and set $\theta$ and $\sigma$ to 0.7 and 0.8 respectively based on a grid search of values in $[0.5, 0.6, 0.7, 0.8, 0.9]$ using 5-fold cross validation.  
We set the maximum number of Gibbs sampling iterations to 1500. 
We compare Context-GPU with the following baselines:

\begin{itemize}
    \item LDA. We use the LDA implementation in  MALLET\footnote{http://mallet.cs.umass.edu} with the default settings and perform hyperparameter optimization every 200 iterations. 
    \item Generalized P\'{o}lya urn (GPU) model \cite{Mimno11}. 
    We implemented this algorithm by modifying the LDA implementation in the MALLET library.
    \item TopicVec \cite{Li16}. 
    We use the available implementation\footnote{https://github.com/askerlee/topicvec} with the default configuration, increasing the maximum iteration number.
    \item TPM \cite{He16Med}. We implemented the Topical Phrase Model which extracts medical topics using both MedTagger and a hierarchy of Pitman-Yor processes. It outperformed other topical phrase extraction models. 
\end{itemize}

\subsection{Topic coherence}
We assess the generated topics by evaluation of their topic coherence. 
We adopt the topic coherence measure proposed in Mimno et al. (\citeyear{Mimno11}), which relies on the co-occurrence statistics collected from the analyzed corpus; this allows us to directly measure the coherence of topics with topical phrases (e.g. \textit{short\_of\_breath}). 

\begin{figure}[!ht]
  \centering
\includegraphics[width=0.48\textwidth]{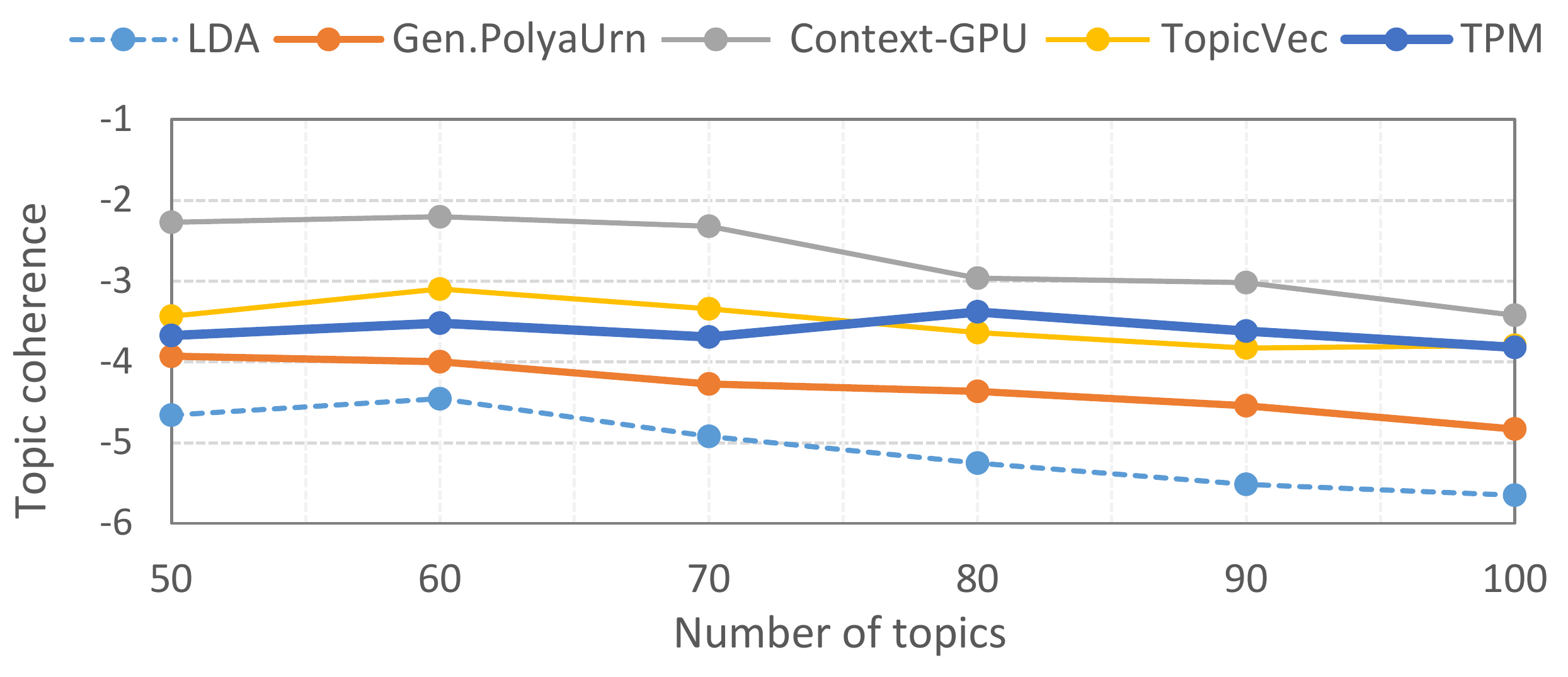}
\caption{Topic coherences vs. number of topics.}
\label{fig:coherences}
\end{figure}

In our evaluations, we compute the topic coherence on the top 10 words/phrases using the implementation provided in the Palmetto library\footnote{https://github.com/dice-group/Palmetto} \cite{Roder15}. 
In Figure \ref{fig:coherences} we report the topic coherence computed by averaging the coherences resulting for each topic. 
A peak of coherence is obtained around 60/70 topics for every model, suggesting a potentially suitable number of topics to discriminate the documents. GPU with only local context incorporated outperforms LDA, but its performance is worse compared to TopicVec or TPM. 
Context-GPU gives superior results over all the baseline models, in particular around 60 and 70 topics. This shows that additionally incorporating global context is essential to achieve better topic coherence results compared to only considering local context. Also, our proposed Context-GPU only involves simple modifications to GPU, but it appears to be more effective than more complicated ways of incorporating word embeddings into topics models (such as TopicVec) or assuming word generation following the HPYP process (such as TPM).

\begin{figure}[h!]
  \centering
\includegraphics[width=0.48\textwidth]{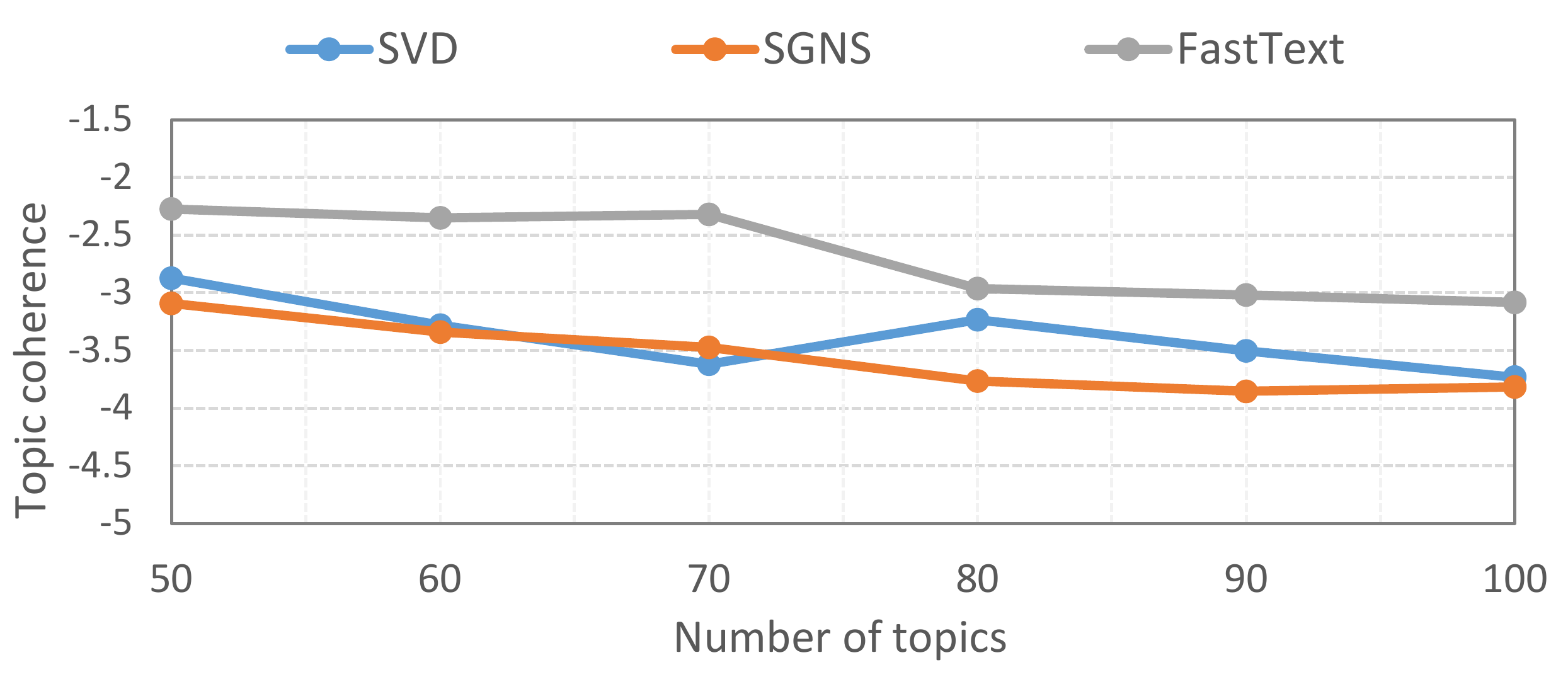}
\caption{Context-GPU with different word/phrase embedding learning methods vs. number of topics.}
\label{fig:embeddings}
\end{figure}

To extract topical phrases from text, we have explored a few different ways in learning word/phrase representations such as learning directly from our data using SVD, training a combined Wikipedia/clinical report data using SGNS or FastText. In Figure \ref{fig:embeddings} we compare these word/phrase embedding learning results over our Context-GPU. 
We can observe that SVD and SGNS perform similarly in most cases and SVD even slightly outperforms SGNS when the topic number is set to 80 or 90. FastText outperforms the other two word/phrase embedding learning methods especially when the topic number is lower than 80. This shows that FastText built on character $n$-grams is more effective in capturing phase sub-structures. 

\begin{table*}[!ht]
\centering
\resizebox{\textwidth}{!}{ 
\begin{tabular}{@{}lllll@{}}
    \toprule
    \textbf{Topic 1}  & \textbf{Topic 2} & \textbf{Topic 3} & \textbf{Topic 4} \\
    \midrule
    \multicolumn{4}{c}{\textbf{TopicVec}}\\
    \midrule
    carotid		    & diuresis							&	dyspnea on exertion		& 			congestive heart failure \\
    coronary artery &		torsemide					&	ejection fraction 		& fibrillation\\
    magnesium		& cardiomyopathy					&	pulmonary  				& ejection fraction\\
    saphenous vein graft	&	shortness of breath		& 	atrial fibrillation  			& insufficiency \\
    potassium chloride	&	torsemide 100 mg			&	diuresed 				& calcium\\
    coronary artery bypass grafting	&	spironolactone 25 mg	&	congestive heart failure 	& intubation\\
    mitral insufficiency	&	diuretic 				&   ischemia				 & thyroid \\
    mitral regurgitation	&	aldactone							&	diabetes mellitus  		& vascular congestion \\
    potassium	&	pleural effusion								&	propafenone				& tricuspid regurgitation\\
    substernal	&	pulmonary edema								&	volume overloaded 		& right knee\\
    \midrule
     \multicolumn{4}{c}{\textbf{Contex-GPU}} \\
    \midrule
  pregnancy			    &	mitral regurgitation 		&	coronary artery disease	 &	congestive heart failure \\
  ultrasound			&	digoxin 					&	cardiac transplant		&	pulmonary edema \\
  postpartum hemorrhage	&	pleural effusion 			&	cardiomyopathy			&	orthopnea \\
  endometrial biopsy  	&	orthopnea		 			&	right coronary artery		&	nonischemic \\
  total abdominal hysterectomy &	dilated cardiomyopathy	& pravachol 20 mg		&	diastolic dysfunction \\
  postpartum			&	plavix 75 mg 					&	paroxysmal atrial fibrillation  	&	cardiomyopathy \\
  vomiting				&	shortness of breath			&   cyclosporine			&	heart failure \\
  salpingo oophorectomy	&	dyspnea on exertion			&   herpes zoster			&	shortness of breath \\
  physical examination	&	tachyarrhythmia 				&	fenofibrate tricor		&	cardiac catheterization \\
  fibroid				&	pulmonary edema 			&	right coronary artery	&	atrial fibrillation \\
    \bottomrule
  \end{tabular}
}
\caption{Topics generated by TopicVec and Context-GPU in 70-topic runs.}
\label{tb:topic_examples}
\end{table*}

\begin{figure}[h!]
  \centering
\includegraphics[width=0.48\textwidth]{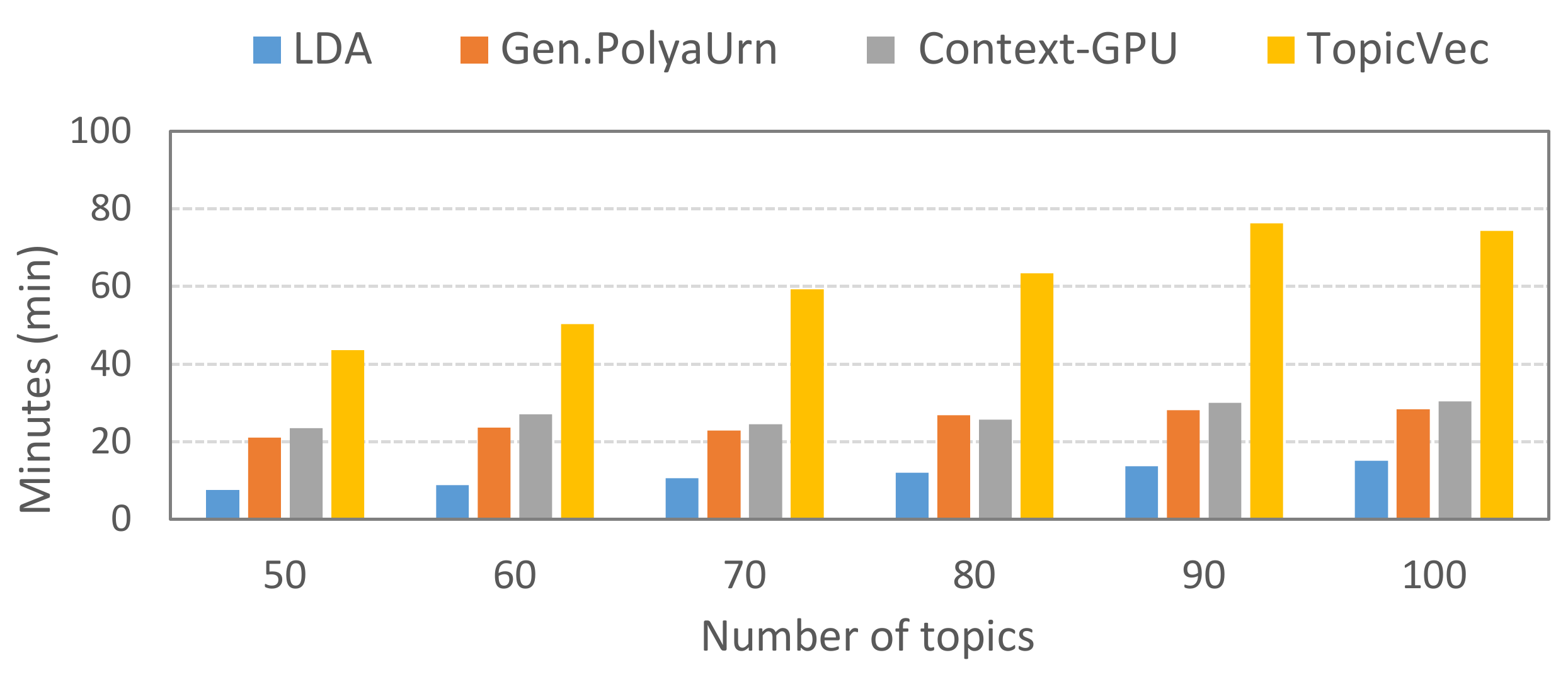}
\caption{Execution time vs. number of topics.}
\label{fig:time}
\end{figure}

Finally, we compare in Figure \ref{fig:time} the execution time needed to train the models, excluding the constant time required by each model to load the embeddings. We did not plot the training time for TPM in the figure as it required significantly more time (over 12 hours) compared to all the other models, showing that modeling phrase generation using HPYP is very expensive. For the remaining models, TopicVec is computationally more complex than the others. Both GPU and Context-GPU have no noticeable difference and they both required three-fold the training time of LDA. Overall, Context-GPU appears to be more effective compared to TopicVec and TPM.

\subsection{Topic Qualitative Assessment}

We report in Table \ref{tb:topic_examples} some topics generated in a 70-topics run. For the sake of brevity, we report only the topics of TopicVec and Context-GPU since TopicVec gives similar coherence scores as TPM but requires significantly less training time compared to TPM.
%
TopicVec inference learns both word and topic embeddings simultaneously. It allows the model to take into account the local context of words, which in turn, alleviates the lack of global statistic for a term. Both the topics of TopicVec and Context-GPU are able to generate topical phrases.  
However, in several topics of Context-GPU, we can distinguish a gradual definition of the analyzed themes, which reflect better semantic coherence. For example, in Topic 4, it can be observed a gradual topic refinement under Context-GPU from the general purpose terms such as \textit{felt} or \textit{insufficiency} to more characterizing words/phrases such as \textit{shortness of breath}, \textit{atrial fibrillation}. In addition, we can observe under the same topic symptom and medication, such as \textit{cardiomyopathy} and \textit{plalix 75 mg}. As a result, the overall expressiveness of topics extracted by the Context-GPU outperforms TopicVec, both thanks to their internal coherence and to the improved expressiveness of the adopted words/phrases.

\section{Conclusion}

We have described a new approach which aims to effectively combine the local and global context of words and phrases. It first detects high reliable phrases and then generates topics using our proposed Context-aware P\'{o}lya urn model. This statistical model combines the word semantic encoded by the context-based and corpus-based embeddings. In particular, we have exploited the LSA and FastText embeddings. The former improved the ties of a word to the corpus themes; the latter allowed a fine-grained use of a word depending on the phrase in which it occurs. An experimental comparison with the state-of-the-art methods has shown an improved coherence of final topics and a decreased computational cost.

\bibliographystyle{aaai}%
\bibliography{topicModels}

\begin{thebibliography}{}

\bibitem[\protect\citeauthoryear{Bergamaschi and Po}{2015}]{Bergamaschi15}
Bergamaschi, S., and Po, L.
\newblock 2015.
\newblock {\em Comparing LDA and LSA Topic Models for Content-Based Movie
  Recommendation Systems}.
\newblock Springer International Publishing.

\bibitem[\protect\citeauthoryear{Blei, Ng, and Jordan}{2003}]{Blei03}
Blei, D.~M.; Ng, A.~Y.; and Jordan, M.~I.
\newblock 2003.
\newblock Latent dirichlet allocation.
\newblock {\em Journal of machine Leraning research} 3:993--1022.

\bibitem[\protect\citeauthoryear{Das, Zaheer, and Dyer}{2015}]{Das15}
Das, R.; Zaheer, M.; and Dyer, C.
\newblock 2015.
\newblock Gaussian lda for topic models with word embeddings.
\newblock In {\em Proc. of the 53rd Annual Meeting of the ACL and the 7th Int.
  Joint Conference on NLP (Volume 1: Long Papers)}.

\bibitem[\protect\citeauthoryear{Deerwester \bgroup et al\mbox.\egroup
  }{1990}]{Deerwester90}
Deerwester, S.; Dumais, S.; Furnas, G.; Landauer, T.; and Harshman, R.
\newblock 1990.
\newblock Indexing by latent semantic analysis.
\newblock {\em Journal of the American Society for Information Science 41}
  391--407.

\bibitem[\protect\citeauthoryear{Dieng \bgroup et al\mbox.\egroup
  }{2017}]{dieng2017topicrnn}
Dieng, A.~B.; Wang, C.; Gao, J.; and Paisley, J.
\newblock 2017.
\newblock Topicrnn: A recurrent neural network with long-range semantic
  dependency.
\newblock In {\em Proceedings of the 5th International Conference on Learning
  Representations (ICLR)}.

\bibitem[\protect\citeauthoryear{Fei, Chen, and Liu}{2014}]{Fei14}
Fei, G.; Chen, Z.; and Liu, B.
\newblock 2014.
\newblock Review topic discovery with phrases using the p{\'{o}}lya urn model.
\newblock In {\em {COLING} 2014, 25th International Conference on Computational
  Linguistics, Dublin, Ireland},  667--676.

\bibitem[\protect\citeauthoryear{He}{2016}]{He16Med}
He, Y.
\newblock 2016.
\newblock Extracting topical phrases from clinical documents.
\newblock In {\em Thirtieth AAAI Conference on Artificial Intelligence,
  AAAI2016}.

\bibitem[\protect\citeauthoryear{Huang and
  Anandkumar}{2016}]{huang2016unsupervised}
Huang, F., and Anandkumar, A.
\newblock 2016.
\newblock Unsupervised learning of word-sequence representations from scratch
  via convolutional tensor decomposition.
\newblock {\em arXiv preprint arXiv:1606.03153}.

\bibitem[\protect\citeauthoryear{Joulin \bgroup et al\mbox.\egroup
  }{2016}]{Joulin16}
Joulin, A.; Grave, E.; Bojanowski, P.; and Mikolov, T.
\newblock 2016.
\newblock Bag of tricks for efficient text classification.
\newblock {\em CoRR} abs/1607.01759.

\bibitem[\protect\citeauthoryear{Lau, Baldwin, and
  Cohn}{2017}]{lau2017topically}
Lau, J.~H.; Baldwin, T.; and Cohn, T.
\newblock 2017.
\newblock Topically driven neural language model.
\newblock In {\em Proceedings of the 55th Annual Meeting of the Association for
  Computational Linguistics (ACL)}.

\bibitem[\protect\citeauthoryear{Le and Mikolov}{2014}]{le2014distributed}
Le, Q., and Mikolov, T.
\newblock 2014.
\newblock Distributed representations of sentences and documents.
\newblock In {\em Proceedings of the 31st International Conference on Machine
  Learning (ICML-14)},  1188--1196.

\bibitem[\protect\citeauthoryear{Levy, Goldberg, and Dagan}{2015}]{Levy15}
Levy, O.; Goldberg, Y.; and Dagan, I.
\newblock 2015.
\newblock Improving distributional similarity with lessons learned from word
  embeddings.
\newblock {\em Transactions of the Association for Computational Linguistics}
  3:211--225.

\bibitem[\protect\citeauthoryear{Li \bgroup et al\mbox.\egroup
  }{2016a}]{Li16SIGIR}
Li, C.; Wang, H.; Zhang, Z.; Sun, A.; and Ma, Z.
\newblock 2016a.
\newblock Topic modeling for short texts with auxiliary word embeddings.
\newblock In {\em Proceedings of the 39th International ACM SIGIR Conference on
  Research and Development in Information Retrieval}, SIGIR '16,  165--174.

\bibitem[\protect\citeauthoryear{Li \bgroup et al\mbox.\egroup }{2016b}]{Li16}
Li, S.; Chua, T.-S.; Zhu, J.; and Miao, C.
\newblock 2016b.
\newblock Generative topic embedding: a continuous representation of documents.
\newblock In {\em Proc. of the 54th Annual Meeting of the ACL (Volume 1: Long
  Papers)},  666--675.

\bibitem[\protect\citeauthoryear{Liu \bgroup et al\mbox.\egroup }{2012}]{liu12}
Liu, H.; Wu, S.~T.; Li, D.; Jonnalagadda, S.; Sohn, S.; Wagholikar, K.; Haug,
  P.~J.; Huff, S.~M.; and Chute, C.~G.
\newblock 2012.
\newblock Towards a semantic lexicon for clinical natural language processing.
\newblock In {\em AMIA Annual Symposium Proceedings}, volume 2012,  568.

\bibitem[\protect\citeauthoryear{Mahmoud}{2008}]{Mahmoud08}
Mahmoud, H.
\newblock 2008.
\newblock {\em Polya Urn Models}.
\newblock Chapman \& Hall/CRC, 1 edition.

\bibitem[\protect\citeauthoryear{Mikolov \bgroup et al\mbox.\egroup
  }{2013a}]{Mikolov13a}
Mikolov, T.; Chen, K.; Corrado, G.; and Dean, J.
\newblock 2013a.
\newblock Efficient estimation of word representations in vector space.
\newblock {\em CoRR 2013}.

\bibitem[\protect\citeauthoryear{Mikolov \bgroup et al\mbox.\egroup
  }{2013b}]{Mikolov13b}
Mikolov, T.; Sutskever, I.; Chen, K.; Corrado, G.; and Dean, J.
\newblock 2013b.
\newblock Distributed representations of words and phrases and their
  compositionality.
\newblock In {\em Proc. of the 26th Int. Conf. on Neural Information Processing
  Systems}, NIPS'13,  3111--3119.

\bibitem[\protect\citeauthoryear{Mimno \bgroup et al\mbox.\egroup
  }{2011}]{Mimno11}
Mimno, D.; Wallach, H.~M.; Talley, E.; Leenders, M.; and McCallum, A.
\newblock 2011.
\newblock Optimizing semantic coherence in topic models.
\newblock In {\em Proceedings of the Conference on Empirical Methods in Natural
  Language Processing}, EMNLP '11,  262--272.

\bibitem[\protect\citeauthoryear{Nguyen \bgroup et al\mbox.\egroup
  }{2015}]{nguyen2015improving}
Nguyen, D.~Q.; Billingsley, R.; Du, L.; and Johnson, M.
\newblock 2015.
\newblock Improving topic models with latent feature word representations.
\newblock {\em Transactions of the Association for Computational Linguistics,
  ACL 2015} 3:299--313.

\bibitem[\protect\citeauthoryear{Rekabsaz \bgroup et al\mbox.\egroup
  }{2017}]{Rekabsaz17}
Rekabsaz, N.; Lupu, M.; Hanbury, A.; and Zamani, H.
\newblock 2017.
\newblock Word embedding causes topic shifting; exploit global context!
\newblock In {\em Proceedings of the 40th International ACM SIGIR Conference on
  Research and Development in Information Retrieval}, SIGIR '17,  1105--1108.

\bibitem[\protect\citeauthoryear{R\"{o}der, Both, and
  Hinneburg}{2015}]{Roder15}
R\"{o}der, M.; Both, A.; and Hinneburg, A.
\newblock 2015.
\newblock Exploring the space of topic coherence measures.
\newblock In {\em Proceedings of the Eighth ACM International Conference on Web
  Search and Data Mining}, WSDM '15,  399--408.

\bibitem[\protect\citeauthoryear{Uzuner \bgroup et al\mbox.\egroup
  }{2010}]{Uzuner10}
Uzuner, O.; Solti, I.; Xia, F.; and Cadag, E.
\newblock 2010.
\newblock Community annotation experiment for ground truth generation for the
  i2b2 medication challenge.
\newblock {\em Journal of the American Medical Informatics Association}
  17(5):519--523.

\bibitem[\protect\citeauthoryear{Yu and Dredze}{2015}]{yu2015learning}
Yu, M., and Dredze, M.
\newblock 2015.
\newblock Learning composition models for phrase embeddings.
\newblock {\em Transactions of the Association for Computational Linguistics,
  ACL 2015} 3:227--242.

\end{thebibliography}

\end{document}